# Visual Place Recognition

Bailu Guo
*ANU College of Engineering and Computer Science*
*Australian National University*
Canberra, Australia
u7233644@anu.edu.au

Boyu Zhao
*ANU College of Engineering and Computer Science*
*Australian National University*
Canberra, Australia
u7237403@anu.edu.au

Zishun Zhou
*ANU College of Engineering and Computer Science*
*Australian National University*
Canberra, Australia
u7211358@anu.edu.au

*Abstract*—Visual position recognition affects the safety and accuracy of automatic driving. To accurately identify the location, this paper studies a visual place recognition algorithm based on HMM filter and HMM smoother. Firstly, we constructed the traffic situations in Canberra city. Then the mathematical models of the HMM filter and HMM smoother were performed. Finally, the vehicle position was predicted based on the algorithms. Experiment results show that HMM smoother is better than HMM filter in terms of prediction accuracy.

## I. INTRODUCTION

Visual place recognition is a well-defined problem: given an image taken at a certain place, people, animals, computers or robots should judge whether the corresponding place of the image has been seen before; If it has been seen, where is the image taken [1]. This technique provides basic position information for automatic driving, and its accuracy directly determines the safety and accuracy of automatic driving. Therefore, the research on visual place recognition is particularly basic and important.

Due to the large amount of data, low latency and the limitation of the hardware in autonomous driving vison-based scenarios, it is required to develop an algorithm that can not only process the image data output by the camera sensor quickly and efficiently but also reduce the storage and computing burden for the hardware chips.

For vision-based location recognition, the sophisticated local-invariant feature extractors, such as Scale-Invariant Feature Transformation and Speed-Up Robust Features, hand-crafted global image descriptors, such as Generalized Search Tress, and the bag-of-visual-words approach were widely used [2]. Recent years, with the development of the deep learning, the convolution neural network is a widely used technique. It offers state-of-the-art performance on many category-level recognition tasks, such as object classification, scene recognition, and image classification.

However, all the techniques above required not only high-performance computing of the hardware but also a huge storage of the memory. It needs to store all the past picture data and compare it with the sensor's current measurements, which will highly increase the cost of the autonomous vehicle.

Markov chain has the characteristics of the current state only depends on the previous state, fast computation, easy deployment and does not require too high performance. Hidden Markov Chains only require the measurements of each current state. This feature can significantly improve the speed of the recognition and highly reduce the amount of data to be processed and stored for the place recognition. In this paper, we proposed two methods based on Hidden Markov Chains.

Filtering is the problem if estimating the current state at this time given the history of the sensor (camera) measurements. Hidden Markov filter obtains the estimated current state based on the past measurements, while hidden Markov Smoother obtains the estimated state based on the measurements before and after it, which are forward pass and backward pass. It is the problem of estimating the state at this time given past, present and future sensor measurements. The key point in the algorithms above is to obtain a reasonable transmission matrix that can compute to estimate the current state of the autonomous driving vehicle. On the basis of the filter, the smoother introduces more state data, which will improve the accuracy of location recognition.

Thus, in this paper, the filter and smoother of Hidden Markov Chains were used in place recognition. It is expected that the algorithms can avoid the hardware performance requirements while maintaining high accuracy. It can improve computing speed and simplify the complexity of location recognition algorithms. This paper is organized as follows. In section 1, we introduced the topic of this paper, system modeling and problem statement is in section 2, the proposed algorithm is described in section 3, detailed simulation result is shown in section 4 and in section 5 summarize this paper.

## II. SYSTEM MODELING

As mentioned above, this paper mainly discusses the problem of re-localization in autonomous driving systems. Different from existing hierarchical clustering-based bag-of-words methods, this paper introduces a directed graph-based Hidden Markov Model. Combined with observation it can relocate vehicle positions. This paper uses a Hidden Markov Model based filter to solve the location. In addition, this paper also uses HMM based smoother to verify the results.

### A. map prior probability

In order to describe the actual position of the vehicle in the map, we establish a position set $V = p_1, p_2, \dots p_i, \dots, p_M$. Where $p_i \in V$ represents a certain position in the map. Let $E_{ij}$ represent the connection from the $p_i$ to the $p_j$, where $i, j \in M$. Let $\Phi_{ij}$ represent the transition probability from position $p_i$ to position $p_j$, that is, the probability that the next status is $p_j$ when the current position is $p_i$, which also can be represented as $P(p_j|p_i)$. Since $\Phi_{ij}$ is expressed as probability, the sum of transition probabilities should be one.

$$\sum_{j}^{j \in s} \Phi_{i,j} = 1 \qquad (1)$$

where $s$ is the position directly connected to the current position. Since the transition probability is position dependent, $E_{ij} \neq E_{ji}$, which is $P(p_j|p_i) \neq P(p_i|p_j)$. From this we can build a directed graph.

$$G(V, E, \Phi)$$

In the graph, $V, E, \Phi$ represent the vertex, edge, and edge weight, respectively, and the specific definitions are as above. As shown in the figure below, this project uses the Canberra city center map as the system map then simulates the status by combining the actual traffic flow data. The red vertex in the figure represents the position $p_i$ in the model, and the blue edge represents that the position $p_i$ is directly connected to $p_j$, that is, $P(p_j|p_i) \neq 0$. We establish different transition probabilities $\Phi_{ij}$ for different edges according to the traffic flow. At vertex such as 1 to 9 on the main road, there will be a higher probability of going straight and a lower probability of stopping. Take node 1 as an example, $\Phi_{1,2} > \Phi_{1,70} > \Phi_{1,1}$. For nodes that are not on the main road, such as node 66, the probability of driving in each direction is roughly the same, and there is also a certain probability of parking. The full transition probability table designed in this paper is attached in the appendix.

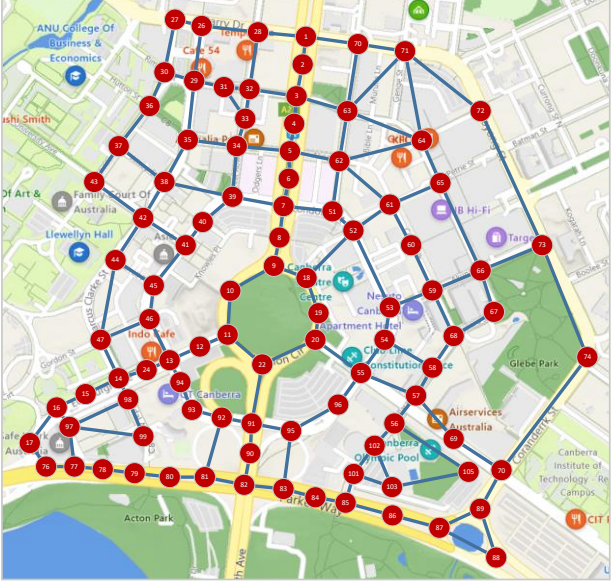

Fig.1 The connection between positions in the simulated system.

### B. vehicle position measurement

Let the observation result of the vehicle at a certain position $x$ be $y_x$, where $y, x \in V$. The vehicle may have different observations at this location, and the probability of observing different positions can be expressed as the following probability.

$P(y = p_j|x = p_i)$, where $\sum_{p_j}^{p_j \in V} P(y = p_j|x = p_i) = 1$.

This project uses designed discrete observation probabilities then add Gaussian noise to simulate real observations. Let the preset discrete observation probability be $P_c$, and the superimposed Gaussian noise be $P_{gaussian}$.

In this project, different observation probabilities are designed according to different road conditions. Among them, the probability between the nodes directly connected to the node is not 0, and the observation probability of the node cannot directly connect to the current node is 0. Since this project believes that the sensor itself has high measurement accuracy, the probability of correct observation is much greater than the probability of incorrect observation, namely.

$$P(y = p_i|x = p_i) \gg P(y \neq p_i|x = p_i)$$

On average $P(y = p_i|x = p_i) \approx 0.7$. The final observation probability of the vehicle still needs to add Gaussian noise on the basis of the discrete probability described above to simulate the noise situation in the observation, then normalize it so that the probability sum is 1. The Gaussian noise superimposed in this project is $\mu = p_i, \delta = 1$. The final observation probability is as follows:

$$P(y = p_j|x = p_i)$$
$$= \left(P_c(y = p_j|x = p_i) + P_{gaussian}(p_j; p_i)\right) \times \delta \quad (2)$$

$$P_{gaussian}(p_j; p_i) = \frac{1}{\sqrt{2\pi}} exp\left(-\frac{(p_j - p_i)^2}{2}\right) \quad (3)$$

where $\delta$ is the normalization coefficient,

$$\delta = 1 + \sum_{p_j}^{p_j \in V} P_{gaussian}(p_j; p_i) \quad (4)$$

### III. PROPOSED INFERENCE ALGORITHM

HMM is an unobservable motion sequence randomly generated by a hidden Markov chain. We use HMM filter and HMM smoother for visual place recognition. In this way, we could use time sequence of the images and the high relationship between the time and position due to the limited movement [3]. The mathematical details of HMM filter and HMM smoother are shown below.

#### A. HMM filter

- The vector $\hat{X}_k \in R^n$ is defined as the filter estimate, which represents the conditional probability mass function of $X_k$ given $y_1, y_2, \ldots, y_k$:

$$\hat{X}_k = \begin{bmatrix} p(X_k = 1|y_1, y_2, \ldots, y_k) \\ p(X_k = 2|y_1, y_2, \ldots, y_k) \\ \vdots \\ p(X_k = n|y_1, y_2, \ldots, y_k) \end{bmatrix} \quad (5)$$

- The matrix $A \in R^{n \times n}$ is defined as the transition probability matrix:

$$A = \begin{bmatrix} p(X_k = 1|X_{k-1} = 1) & \cdots & p(X_k = 1|X_{k-1} = n) \\ \vdots & \ddots & \vdots \\ p(X_k = n|X_{k-1} = 1) & \cdots & p(X_k = n|X_{k-1} = n) \end{bmatrix} \quad (6)$$

- Diagonal matrix $B(y_k) \in R^{n \times n}$ is defined with likelihoods for measurement $y_k$ on its diagonal:

$$B(y_k) = \begin{bmatrix} p(y_k|X_{k-1} = 1) & \cdots & 0 \\ \vdots & \ddots & \vdots \\ 0 & \cdots & p(y_k|X_{k-1} = n) \end{bmatrix} \quad (7)$$

- Then we could written the HMM filter as follows:

$$\hat{X}_k = N_k^{-1} B(y_k) A \hat{X}_{k-1} \quad (8)$$

where $A\hat{X}_{k-1}$ represents the "prediction step", $N_k^{-1}B(y_k)$ represents the "update step", and $N_k \in$

$R$ normalizes to ensure $\hat{X}_k$ is a probability mass function, i.e.,

$$N_k = \sum_{i=1}^{n} V(i) \text{ with } V = B(y_k) A \hat{X}_{k-1}$$

### B. HMM smoother

The first step for HMM smoother called Forward Pass. In this step, we have to calculate the unnormalized filter estimate. The details are as follows.

- The vector $\alpha_k \in R^n$ is defined as the unnormalized filter estimate:

$$\alpha_k = \begin{bmatrix} p(X_k = 1, y_1, y_2, \ldots, y_k) \\ p(X_k = 2, y_1, y_2, \ldots, y_k) \\ \vdots \\ p(X_k = n, y_1, y_2, \ldots, y_k) \end{bmatrix} \quad (9)$$

- When k change from 1 to T, compute the unnormalized filter estimate:

$$\alpha_k = B(y_k) A \alpha_{k-1} \quad (10)$$

with $\alpha_0 = \pi_0$

The second step called Backward Pass. In this step, we have to calculate the unnormalized backward filter estimate. The details are as follows.

- The vector $\beta_k \in R^n$ is defined as the unnormalized backward filter estimate:

$$\beta_k = \begin{bmatrix} p(X_k = 1, y_{k+1}, y_{k+2}, \ldots, y_T) \\ p(X_k = 2, y_{k+1}, y_{k+2}, \ldots, y_T) \\ \vdots \\ p(X_k = n, y_{k+1}, y_{k+2}, \ldots, y_T) \end{bmatrix} \quad (11)$$

- When k change from T to 1, compute the unnormalized backward filter estimate:

$$\beta_{k-1} = A' B(y_k) \beta_k \quad (12)$$

where $\beta_T = 1$ representing the vector of ones, and $A'$ represents the transpose of A.

The final step called Multiply and Normalize. In this step, we have to calculate the smoother estimate, which represents the final result. The details are as follows.

- The vector $\gamma_k \in R^n$ is defined as the smoother estimate:

$$\gamma_k = \begin{bmatrix} p(X_k = 1 | y_1, y_2, \ldots, y_T) \\ p(X_k = 2 | y_1, y_2, \ldots, y_T) \\ \vdots \\ p(X_k = n | y_1, y_2, \ldots, y_T) \end{bmatrix} \quad (13)$$

- When k change from 0 to T, computer and normalize via elementwise multiplication:

$$\gamma_k(x_k) = \frac{\alpha_k(x_k) \beta_k(x_k)}{\sum \alpha_k(x_k) \beta_k(x_k)} \quad (14)$$

for all $x_k \in \{1, 2, \ldots, n\}$ [4].

## IV. SIMULATION RESULTS

We selected 105 traffic intersections in Canberra central, and found the current location of cars through HMM filter and HMM smoother model. To make comparison, we set the length of the state sequence to be generated to 50. It was required to design an evaluation method. In this simulation, we used accuracy.

$$accuracy = \frac{Number\ of\ correct\ estimated\ states}{Number\ of\ all\ states} \quad (15)$$

The simulation results are shown below.

### A. Initial state = 5

We started experiment from setting initial state as 5 and did the following experiments.

*1) Initial state=5, Sigma = 1*

Firstly, we set the covariance of Gaussian Variance as 1. The HMM filter result can be seen in the image below.

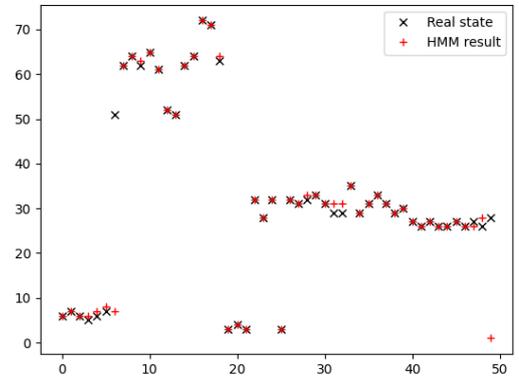

Fig.2 Filter simulation: Initial state=5, sigma=1

The accuracy of HMM filter is 0.76.

The result of HMM smoother can be seen in the image below.

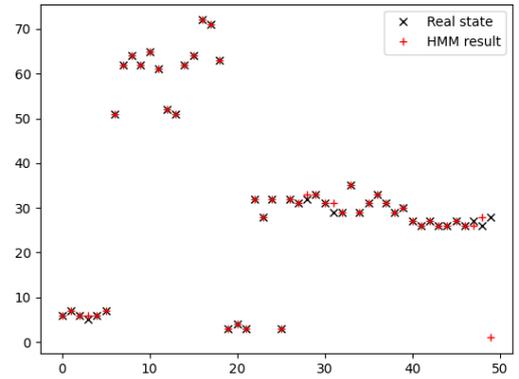

Fig.3 Smoother simulation: Initial state=5, sigma=1

The accuracy of HMM Smoother is 0.88.

The image below shows the performance of filter and smoother when initial state is 5 and sigma is 1.

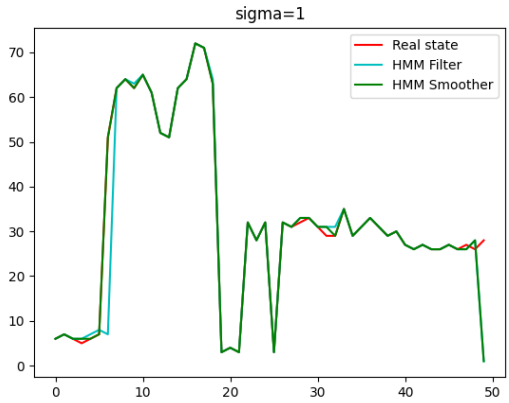

Fig.4 Comparison: Initial state=5, sigma=1

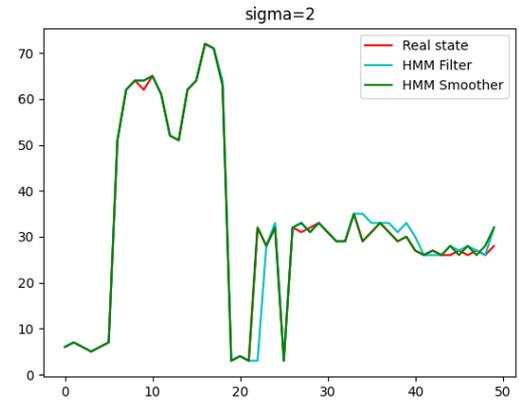

Fig.7 Comparison: Initial state=5, sigma=1

*2) Initial state=5, Sigma = 2*

When setting the covariance of Gaussian noise is 2, the result of HMM filter can be seen in the image below.

B. *Initial state = 90*

Then we set the initial state as 90 Sigma = 1
Similarly, we set the covariance of Gaussian Variance as 1. The HMM filter result can be seen in the image below.

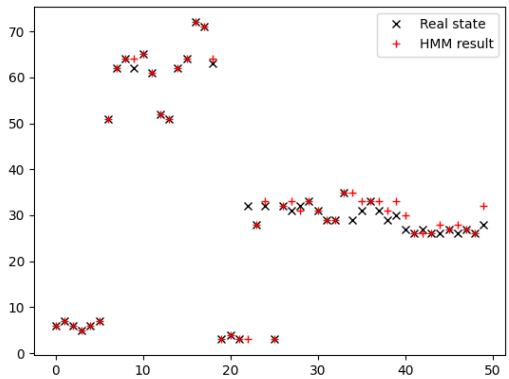

Fig.5 Filter simulation: Initial state=5, sigma=2

The accuracy of HMM filter when sigma equals 2 is 0.68.
The result of HMM smoother can be seen in the image below.

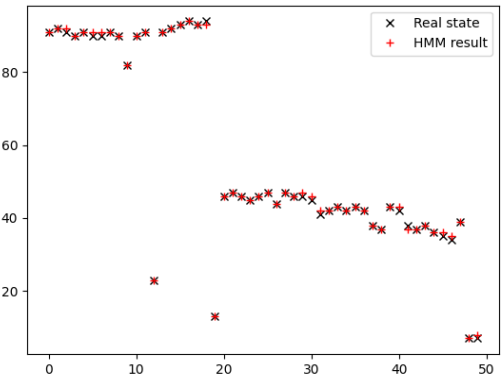

Fig.8 Filter simulation: Initial state=90, sigma=1

The accuracy of HMM filter in this assumption is 0.76.
The result of HMM smoother can be seen in below.

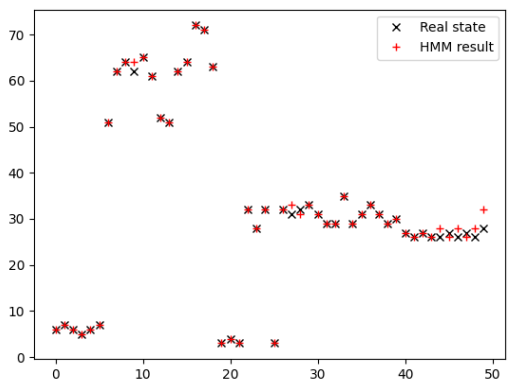

Fig.6 Smoother simulation: Initial state=5, sigma=2

The accuracy of HMM Smoother is 0.82.
The image below shows the performance of filter and smoother when initial state is 5 and sigma is 2.

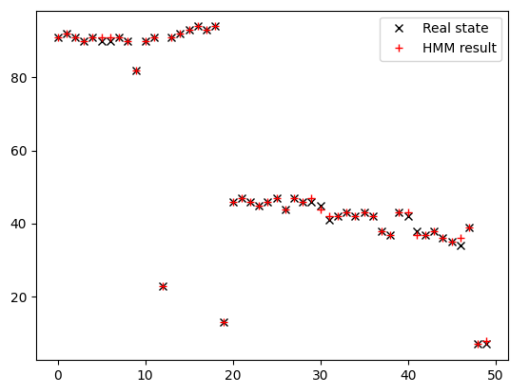

Fig.9 Smoother simulation: Initial state=90, sigma=1

The accuracy of HMM filter in this assumption is 0.82.
The image below shows the performance of filter and smoother when initial state is 90 and sigma is 1.

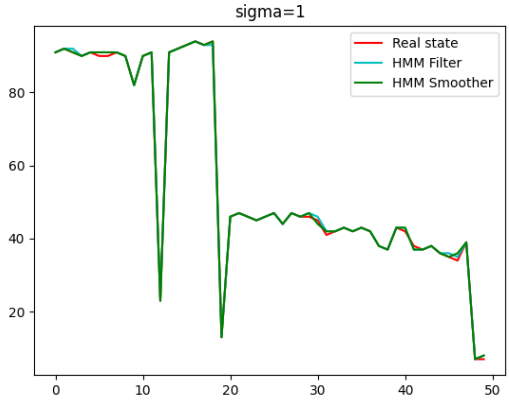

Fig.10 Comparison: Initial state=90, sigma=1

*C. Summary*

We record the accuracy in the figure above in the following table.

TABLE 1. THE ACCURACY OF HMM FILTER AND HMM SMOOTHER

|  | **HMM filter accuracy** | **HMM smoother accuracy** |
| --- | --- | --- |
| Initial state=5 Sigma = 1 | 76% | 88% |
| Initial state=5 Sigma = 2 | 68% | 82% |
| Initial state=90 Sigma = 1 | 76% | 82% |

It could be drawn the following conclusions.

Firstly, it can be found that at the beginning, both the filter and smoother show a good performance. Then it can be seen that the smoother validates a more accurate estimations of the state. We could see that the error between the true and estimated state of the HMM smoother is less than the error of the HMM filter. Especially at the end of the states, the results of the smoother are more consistent with the real state.

Due to the HMM smoother considered not only the past measurements of the autonomous vehicle, but also the future states of the vehicle. In other words, through backward pass, more knowledge is considered into the model when estimating the current state, which will lead a better estimation, while the filter can only estimate the state by using the past knowledge. Thus, it is reasonable that as the time went by, the advantages of the smoother over the filter become more and more obvious.

However, the smoother will meet more limitations in practice. In practical scenarios, it is difficult to obtain a large number of subsequent measurements of the state to be estimated. Thus, to use the smoother in practice, it is required more sensors on the vehicle and a good fusion of the sensors.

Secondly, the added Gaussian noise when generated measurements have also made affects in HMM filter and HMM smoother. It could be found that the higher the variance of the noise, the worse the results of the filter and smoother.

Thirdly, for the states where the mismatch happened, the confusion matrix affected the HMM filter and smoother. In practice, it is the accuracy of the sensor. Due to the errors in sensors when capture the places, the algorithm cannot obtain absolutely accurate measurements as the input, which will lead mismatches in the simulations results. What's more, Bayesian algorithm is based on possibilities. If the prior belief or likelihood are inaccurate, the result will be bound to a mismatch. For this point, the algorithm like HMM smoother that introduces more knowledge will have an even greater advantage. Especially in the use of autonomous driving which requires extreme security and reliability, HMM smoother will be a better choice.

V. CONCLUSION

In this paper, HMM filter and HMM smoother were used for visual position recognition. By comparing the state observation results under different initial positions, we found that the HMM smoother had better performance. By comparing different gaussian noise, we found that the higher the noise variance, the worse the results of filter and smoother. Besides, after the theoretical analysis, we thought that the errors which led to the difference between the real states and the simulation results are the accuracy of the sensor. Thus, In order to improve the accuracy of position recognition, we put forward the following conclusions: firstly, HMM smoother performs a better performance with more limitations. Secondly, the sensors which used to do measurements played an important role in predictions. Improving the performance of the sensors will be an essential solution if the prediction algorithm doesn't work well.

APPENDIX

A. *The transition probability table*

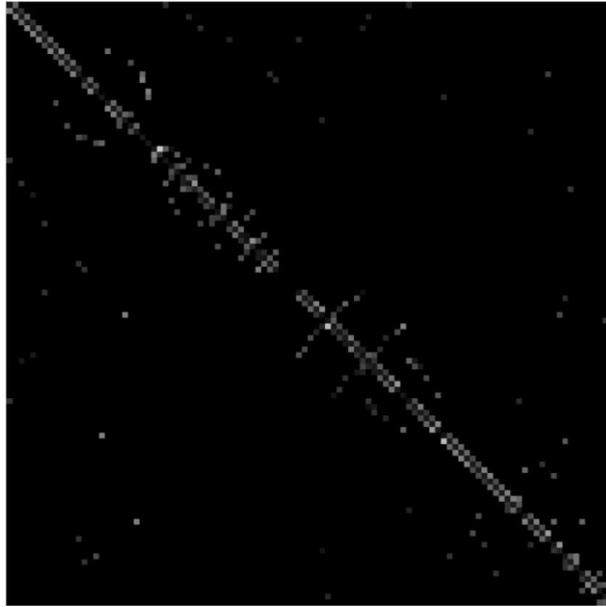

**Figure A.** the full transition matrix, since the matrix has 105 cols and 105 rows, it can only be presented as a figure. This figure has 105x105 pixels, each pixel represents one of the elements in the matrix. The brightness of pixels represents the probability of that element. The lighter pixel will have a larger probability. The sum of each cols are equal to one.

B. *The observation matrix*

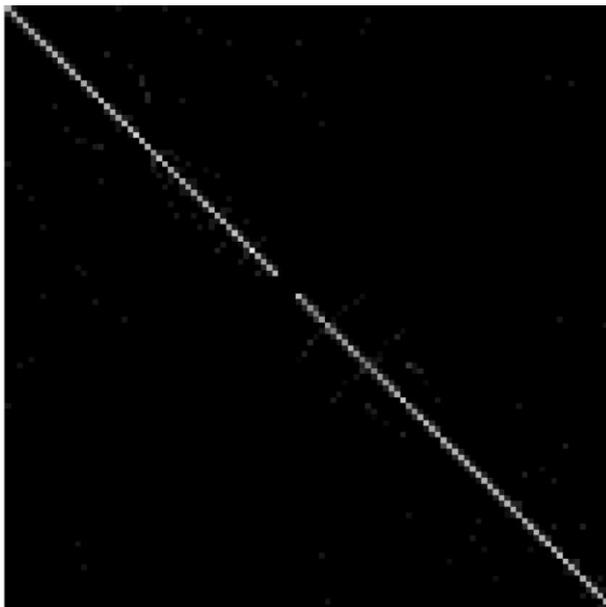

**Figure B** The full observation matrix. This is a 105x105 figure, where each pixel represents one of the elements in observation matrix, the definition of brightness is the same as transition matrix figure. The average value of diagonal elements is 0.7.